\documentclass[11pt]{article}
\usepackage{coling2020}
\usepackage{times}
\usepackage{url}
\usepackage{latexsym}
\usepackage{booktabs}

\usepackage{todonotes}

\usepackage{tcolorbox}

\colingfinalcopy 

\title{
UHH-LT at SemEval-2020 Task 12: 
Fine-Tuning of Pre-Trained Transformer Networks 
for Offensive Language Detection
\vspace{1cm}
}

\author{Gregor Wiedemann \\\And
  Seid Muhie Yimam \\
  Language Technology Group\\ 
  Department of Informatics\\ 
  University of Hamburg, Germany\\
  {\tt \{gwiedemann, yimam, biemann\}@informatik.uni-hamburg.de} \\\And
  Chris Biemann}

\date{}

\begin{document}
\maketitle
\begin{abstract}
  Fine-tuning of pre-trained transformer networks such as BERT yield state-of-the-art results for text classification tasks.
  Typically, fine-tuning is performed on task-specific training datasets in a supervised manner.
  One can also fine-tune in unsupervised manner beforehand by further pre-training the masked language modeling (MLM) task.
  Hereby, in-domain data for unsupervised MLM resembling the actual classification target dataset allows for domain adaptation of the model.
  In this paper, we compare current pre-trained transformer networks with and without MLM fine-tuning on their performance for offensive language detection. 
  Our MLM fine-tuned RoBERTa-based classifier officially ranks 1st in the SemEval 2020 Shared Task~12 for the English language. Further experiments with the ALBERT model even surpass this result.
\end{abstract}

\section{Offensive Language Detection}
\label{introduction}

The automatic detection of hate-speech, cyber-bullying, or aggressive and offensive language became a vividly studied task in natural language processing (NLP) in recent years. 
The offensive language detection (OLD) shared Task~6 of 2019's \textit{International Workshop on Semantic Evaluation} (SemEval) \cite{offenseval} attracted submissions from more than 100 teams. 
The \textit{Offensive Language Identification Dataset} (OLID) used in this shared task comprises three hierarchical classification sub-tasks: A) offensive language detection, B) categorization of offensive language, and C) offensive language target identification \cite{zampieri2019predicting}. 
For Task A, 14,100 Twitter tweets were manually labeled as either \textit{offensive} (OFF) or \textit{not offensive} (NOT). 
Task B distinguishes the 4,640 offensive tweets from task A into \textit{targeted} (TIN) or general, \textit{untargeted} (UNT) offensive language. 
Task C, finally, separates targeted insults into the three categories: groups (GRP), individuals (IND), and others (OTH).

OffensEval 2020, the SemEval~2020 \textit{Offensive Language Detection Shared Task}, does not provide an own manually labeled training set for the English language \cite{zampieri-etal-2020-semeval}. 
Instead, a large `weakly labeled' dataset was published by the organizers, containing roughly nine million tweets. 
Each tweet has been automatically classified by an ensemble of five different supervised classifiers trained on the OLID dataset. 
The weakly labeled dataset contains the raw tweet (with user mentions replaced by a special `USER' token) along with the average label probability and the variance of the five classifier predictions.
Since there is no way that such weak labels themselves carry more useful information to a machine learning system than the original dataset on which the five classifiers were trained, we decided not to use any of the weakly labeled information. 
Instead, for our classification systems, we rely on the 2019 OLID dataset only.
However, the OffensEval 2020 dataset is an ample source to build models using unsupervised learning, particularly for domain-adaptation of a pre-trained language model such as BERT \cite{devlin.2019} or its successors which are based on the transformer neural network architecture. 
Unfortunately, training a transformer-based language model in an unsupervised manner is incredibly resource-consuming, making it impractical to learn from large datasets without access to larger GPU clusters or TPU hardware.
Regarding this, the contribution of our paper is two-fold:
\begin{enumerate}
    \item We evaluate to what extent different pre-trained transformer-based neural network models can be fine-tuned to detect offensive language and its sub-categories. An ensemble based on the ALBERT \cite{lan2019albert} model achieves the best overall performance.
    \item We study how an additional fine-tuning step with masked language modeling (MLM) of the best individual model RoBERTa \cite{liu2019roberta} conducted on in-domain data affects the model performance. An ensemble of models trained with this strategy was submitted as our official contribution to the OffensEval 2020 shared task for the English language and achieved first place in the competition.
\end{enumerate}

\section{Related Work}
\label{relworks}

\paragraph{Offensive language detection:} Nowadays, a number of public datasets are available to train machine classifiers for detecting English offensive language. Unfortunately, underlying data sources, category definitions, data sampling strategies, and annotation guidelines differ to a large extent between these datasets. 
Hence, results of different datasets are hardly comparable, and training sets usually cannot be combined to obtain a more robust classification system. 
\newcite{schmidt2017survey}, and  \newcite{fortuna2018survey} conducted insightful surveys on this rapidly growing field.
\newcite{mandl2019overview}, \newcite{stru2019germeval}, and \newcite{basile2019semeval} recently organized shared tasks on the topic.
Although winning systems can achieve striking prediction accuracy, OLD is far from being a solved problem. Prediction performance usually drops severely if the target data comprises different characteristics than the training data. \newcite{grondal2018adversarial}, for instance, show that many machine learning architectures can be fooled easily just by adding the word ``love'' to an offensive tweet to make it appear as non-offensive. \newcite{stru2019germeval} highlight that linguistic information alone is not enough in many cases to decide whether a tweet is hateful or not. Also context information, e.g. about tweeting users themselves \cite{ribeiro2018users}, or mentioned users in tweets \cite{wiedemann2018germeval} can be a useful feature for automatic OLD.

\paragraph{Pre-trained language models for text classification:} 
Transfer learning with deep neural networks, in general, has proven to be superior over supervised learning for text classification, especially for small training data situations.
This is illustrated exemplarily in our last year's approach \cite{wiedemann2019semeval} to the OLD SemEval shared task which employed unsupervised pre-training of a recurrent neural network architecture with a topic cluster prediction task.
Practically all winners of the aforementioned shared task competitions employ some form of a fine-tuned bidirectional transformer-based language model, a neural network architecture for which \newcite{devlin.2019} published with BERT the seminal work.
This architecture has been proven highly successful for transfer learning. A base model is pre-trained with a MLM task and a next-sentence prediction (NSP) task in an unsupervised manner on very large datasets. 
The knowledge about language regularities and semantic coherence encoded in the network during this step can then be employed successfully in later training steps of fine-tuning the network weights to the actual classification task. 
For instance, \newcite{liu-etal-2019-nuli} fine-tuned the pre-trained BERT model winning the 2019 SemEval OLD shared task. 
Also \newcite{Mozafari2019}, and \newcite{risch-etal-2019-hpidedis} used it successfully for offensive language and hate speech detection.
\newcite{sun2019bertft} test a wide range of BERT fine-tuning methods for text classification and develop best practice recommendations.
Since BERT, a number of successor models improving the network architecture, the pre-training strategy, or the pre-training dataset have been published. A selection of these models will be evaluated in Section~\ref{sec:method}.

\section{Fine-tuning Transformer Networks}
\label{sec:method}

We investigate two questions regarding the fine-tuning of pre-trained transformer networks for OLD. First, which pre-trained model performs best on the 2020 OLD shared task? Second, we investigate how much language model fine-tuning on in-domain data prior to classification fine-tuning improves the performance of the best model.

\subsection{Model Selection of Transformer Networks}
\label{subsec:ft}

As we have indicated in Section \ref{relworks}, transformer networks have been successfully employed for several text classification tasks. We test the following transformer-based pre-trained models for the OffensEval 2020 OLD shared task.

\paragraph{BERT -- Bidirectional Encoder Representations from Transformers:} this seminal transformer- based language model employs an attention mechanism that enables to learn contextual relations between (sub-)words in a text sequence \cite{devlin.2019}. BERT uses two training strategies: 1)~MLM where 15 \% of the tokens in a sequence are replaced (masked) for which the model learns to predict the original tokens, and 2)~NSP where the model receives pairs of sentences as input and learns to predict whether or not the second sentence is a successor of the first one in their original document context. 

\paragraph{RoBERTa -- A Robustly Optimized BERT Pretraining Approach:} this is a replication of BERT developed by Facebook \cite{liu2019roberta} with the following modifications 1) training the model longer with bigger batches as well as more and cleaner data, 2) discard the NSP objective, 3) training on longer sequences, and 4) dynamically change the masking patterns, e.g. taking care of masking complete multi-word units. RoBERTa outperformed BERT on most tasks of the GLUE NLP benchmark (ibid.).

\paragraph{XLM-RoBERTa -- XLM-R:} this is a cross-lingual version of RoBERTa which is trained on several languages at once \cite{conneau2019xlm-roberta}. The model itself is equivalent to RoBERTa, but the training data consists of texts from more than 100 languages filtered from the CommonCrawl\footnote{\url{https://commoncrawl.org}} dataset.  

\paragraph{ALBERT -- A Lite BERT for Self-supervised Learning of Language Representations:} this is a modification on BERT especially to mitigate memory limitations and training time issues \cite{lan2019albert}. The main contributions that ALBERT makes over the design choices of BERT are 1) decomposing the embedding parameters into smaller matrices that will be projected to the hidden space separately, 2) share parameters across layers to improve or stabilize the learned parameters, and 3) inter-sentence coherence loss, which is based on sentence order prediction (SOP), in contrast to BERT's simpler NSP objective. 

\subsection{Masked Language Model Fine-tuning}
\label{subsec:mlm}

\newcite{sun2019bertft} showed that further pre-training of BERT with the masked language model task can improve later results of supervised task-specific fine-tuning. The authors tested \textit{within-task}, \textit{in-domain} and \textit{cross-domain} further pre-training. Evaluations show that the first strategy is susceptible to overfit the training set and, thus, may harm classification performance. The last strategy does not help since BERT is already trained on general-domain data. In-domain further pre-training, however, helps to improve later classification performance if there is a substantial overlap in language characteristics between further pre-training data and supervised training data.

The `weakly labeled' dataset of the 2020 OLD shared task most certainly is a valuable dataset for further in-domain pre-training. However, with ca. 9 million tweets it is also rather large. Pre-training on the complete dataset is not possible regarding our hardware limitations.\footnote{MLM of the RoBERTa-large model with the full dataset on a single GPU with 12 GB RAM would take estimated 40 days. However, due to increasing memory consumption of the Adam optimizer during training, the process will stop unfinished way earlier due to a memory exception.} Therefore, we conduct MLM pre-training only on a small sample of the original data. We strip URLs and user mentions from tweets, remove duplicates and, finally, randomly sample 5 \% of the original dataset size, i.e. 436.123 tweets for further pre-training. We further pre-train the presumably best model \textit{RoBERTa-large} \cite{liu2019roberta} (cp. Section~\ref{sec:results}) for one epoch (batch size 4, and learning rate 2e-5).

\subsection{Ensembling}

For our official OffensEval 2020 test set submission as team \textit{UHH-LT}, we aggregated predictions from classifiers with different ensemble approaches.

\paragraph{Ensemble of model variants:} 
We fine-tuned different transformer models with the OffensEval 2019 training data using the corresponding test data for validation. The following models were tested: BERT-base and BERT-large (uncased), RoBERTa-base and RoBERTa-large, XLM-RoBERTa, and four different  ALBERT models (large-v1, large-v2, xxlarge-v1, and xxlarge-v2). Each model was fine-tuned for 6 epochs with a learning rate of 5e-6,  maximum sequence length of 128, and batch size 4. After each epoch, the model was evaluated on the validation set. The best performing epoch was saved for the ensembling. We tested two ensemble approaches: 1) majority vote from all models, and 2) majority vote from one model type but with different parameter sizes such as BERT-base and BERT-large. 

\paragraph{MLM RoBERTa ensemble:} To be able to learn from the entire 2019 OLID dataset (training and test set), as well as to smooth instabilities of predictions originating from random effects during model training, we also aggregated predictions using 10-fold cross-validation. For this, the further MLM pre-trained RoBERTa-large model is fine-tuned 10 times, each time with 90 \% of the OLID data for training and the remaining 10 \% as validation set. The best model after 6 epochs of training with learning rate 5e-6 and batch size 8 is used to predict the OLD 2020 test data. The final predictions for submission were obtained via majority vote on the 10 predictions per test data instance.

\section{Results}
\label{sec:results}

\begin{table}[]
\centering
\begin{tabular}{l|lll|lll|ll}
\hline
                           & \multicolumn{3}{l|}{\textbf{NOT}}      & \multicolumn{3}{l|}{\textbf{OFF}}      &                   &                   \\ \hline
\textbf{Model}             & \textbf{P} & \textbf{R} & \textbf{F1} & \textbf{P} & \textbf{R} & \textbf{F1} & \textbf{Macro F1} & \textbf{Acc.} \\\hline
\multicolumn{9}{c}{Baselines} \\ \hline
\textbf{All NOT}           & 72.21  & 100.00       & 41.93        & -        & 0.00     & 0.00       & 41.93     & 72.21             \\
\textbf{All OFF}           & -      & 0.00         & 0.00         & 27.78    & 100.00   & 43.49      & 21.74     & 27.79                   \\ \hline
\multicolumn{9}{c}{Single pre-trained transformer models} \\ \hline
\textbf{BERT-base}         & 99.06       &       90.2       &       94.42       &       79.34       &       97.78       &       87.60      &       91.01      &       92.31       \\
\textbf{BERT-large}        & \underline{99.65}       &       90.35       &       94.77       &       79.81       &       \underline{99.17}       &       88.44      &       91.60      &       92.80  \\
\textbf{RoBERTa-base}      & 99.45       &       90.70       &       94.88       &       80.33       &       98.70       &       88.57      &       91.73      &       92.93       \\
\textbf{RoBERTa-large}     & 99.53       &       90.92       &       \underline{95.03}       &       80.73       &       98.89       &       \underline{88.89}      &       \underline{91.96}      &       \underline{93.13}       \\ 
\textbf{XLM-RoBERTa}       & 99.03       &       91.31       &       95.01       &       81.22       &       97.69       &       88.69      &       91.85      &       93.08       \\
\textbf{ALBERT-large-v1}   & 98.87       &       90.24       &       94.36       &       79.32       &       97.31       &       87.40      &       90.88      &       92.20       \\
\textbf{ALBERT-large-v2}   & 98.87       &       90.20       &       94.34       &       79.26       &       97.31       &       87.36      &       90.85      &       92.18       \\
\textbf{ALBERT-xxlarge-v1} & 98.35       &       91.09       &       94.58       &       80.57       &       96.02       &       87.62      &       91.10      &       92.46       \\
\textbf{ALBERT-xxlarge-v2} & 98.47       &       \underline{91.73}       &       94.98       &       \underline{81.76}       &       96.30       &       88.44      &       91.71      &       93.00       \\ \hline
\multicolumn{9}{c}{Ensembles of pre-trained transformer models} \\ \hline
\textbf{All}      & \textbf{99.65}       &       90.95       &       95.10       &       80.83       &       \textbf{99.17}       &       89.06      &       92.08      &       93.23       \\
\textbf{BERT}     & 99.42       &       91.16       &       95.11       &       81.11       &       98.61       &       89.01      &       92.06      &       93.23       \\
\textbf{RoBERTa}  & 99.57       &       90.84       &       95.01       &       80.62       &       98.98       &       88.86      &       91.93      &       93.11       \\
\textbf{ALBERT-all}         &      98.23       &       \textbf{92.66}       &       \textbf{95.36}       &       \textbf{83.37}       &       95.65       &       89.00      &       92.23      &       \textbf{93.49}       \\ 
\textbf{ALBERT-xxlarge}         &     98.70       &       92.16       &       95.32       &       82.62       &       96.85       &       \textbf{89.17}      &       \textbf{92.25}      &       93.47       \\ \hline
\end{tabular}
\caption{Performance (in \%) of baselines, single models, and ensemble models on the OLID test set.
}
\label{tab:result_model_selection}
\end{table}

Table~\ref{tab:result_model_selection} shows results of binary offensive language detection for a naive baseline (assuming all tweets as either offensive or not), as well as for the individual fine-tuned transformer models and their corresponding ensembles. All transformer models largely outperform the naïve baseline, some of them (e.g. XLM-RoBERTa) even outperform most of the other system submissions in the competition.\footnote{\url{https://sites.google.com/site/offensevalsharedtask/results-and-paper-submission}} 

Our best individual model is \textit{RoBERTa-large} with an F1-score of 91.96 \%. Hence, we select this model as the basis for further MLM pre-training. From Table~\ref{tab:result_mlm_selection}, we can see that the MLM fine-tuned RoBERTa model achieved consistently better results in terms of Macro F1 than the single pre-trained transformer models (to lower random effects of neural network training, the table shows average values of 10 runs).

Regarding the ensembles of model variants, we see in Table~\ref{tab:result_model_selection} that all approaches consistently perform better than the individual models. Here, the ensemble averaging the predictions from the two \textit{ALBERT-xxlarge} models performed best with an F1-score of 92.25 \%.

For the OffensEval 2020 Shared Task, we decided to submit the results form the MLM pre-trained RoBERTa ensemble.\footnote{Of course, during the submission phase of the shared task, the test set labels were not available. We, thus, based our decision for this specific model on its performance on last year's OLID test set.}
Table~\ref{tab:result_oe2020} presents the official results of our system in the sub-tasks A, B, and C together with their ranks achieved in the competition. While our ensemble reaches the top rank for task A, there are a handful of competing approaches achieving better results for offensive language categorization (B) and  target identification (C). 
The post-submission experiments on the official test set as presented in this paper (cp. Table~\ref{tab:result_model_selection}) show that the ALBERT-based ensembles would have even beat this first ranked submission.
However, the MLM fine-tuned RoBERTa model was considerably more successful on task C than the best ALBERT-xxlarge ensemble, especially to detect the ``OTH'' class.
\begin{table}[]
\centering
\begin{tabular}{l|lll|lll|ll}
\hline
                           & \multicolumn{3}{l|}{\textbf{NOT}}      & \multicolumn{3}{l|}{\textbf{OFF}}      &                   &                   \\ \hline
\textbf{Model}             & \textbf{P} & \textbf{R} & \textbf{F1} & \textbf{P} & \textbf{R} & \textbf{F1} & \textbf{Macro F1} & \textbf{Acc.} \\\hline
\textbf{RoBERTa-large}            & 98.96      & 91.49      & 95.07       & 81.54      & 97.49      & 88.76       & 91.93      & 93.15             \\
\textbf{RoBERTa-large MLM-ft}     & \textbf{99.15}      & \textbf{91.53}       & \textbf{95.18}       & \textbf{81.66}       & \textbf{97.96}      & \textbf{89.06}       & \textbf{92.12}             & \textbf{93.31}  \\ \hline
\end{tabular}
\caption{Performance (in \%) of MLM fine-tuned models on the OLID test set (average of 10 runs). 
}
\label{tab:result_mlm_selection}
\end{table}
\begin{table}
\centering
\begin{tabular}{l|ll}
\hline
Team                 & \multicolumn{2}{l}{\textbf{UHH-LT}}  \\ \hline
                     & Macro F1 (\%) & Rank  \\ \hline
Task A               & 92.04    &  1 out of 84  \\
Task B               & 65.98    &  6 out of 42  \\
Task C               & 66.83    &  3 out of 38  \\ \hline
\end{tabular}
\caption{Official results of our OffenseEval 2020 test set submissions for tasks A, B, and C (\textit{RoBERTa-large MLM test} ensemble, cp. Table~\ref{tab:result_mlm_selection}).}
\label{tab:result_oe2020}
\end{table}

\begin{figure}[h!]
  \centering
\includegraphics[width=5cm, height=3.91cm]{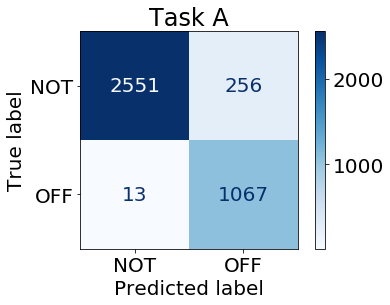}
\includegraphics[width=5cm, height=3.91cm]{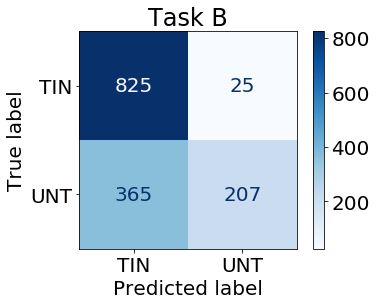}
\includegraphics[width=5cm, height=3.91cm]{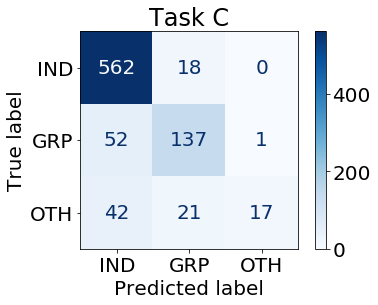}
  \caption{Confusion matrices from the submitted UHH-LT ensemble predictions.}
  \label{fig:confusion_matrix}
\end{figure}

Figure \ref{fig:confusion_matrix} shows the corresponding confusion matrices for the submitted predictions. One observation that can be revealed from the matrices is that the predictions for Tasks B and C are considerably biased towards the majority class. For Task A, however, we see more false positive cases for the offensive class which is underrepresented in the training data.
A qualitative look into a sample of false predictions (cp. Fig.~\ref{fig:examples}) reveals tweets wrongly predicted as offensive (false positives), some of which seem inconsistently annotated in the gold standard. Parts of expressions in Examples 1, and 2 qualified as offensive in many other training examples. Examples 3, and 4 contain some negative words that may have triggered a higher offensiveness prediction score. For the false negative samples, it is not really obvious why the models missed the correct class, since except for example~6, they actually contain strong offensive vocabulary.
\begin{figure}[h!]
  \centering
  \small
{\footnotesize
\begin{tcolorbox}
\begin{enumerate}
	\item FP: \emph{@USER the gov and nyc mayor are the biggest joke except maybe for the idiots who elected them}
	\item FP: \emph{@USER What the fuck}
	\item FP: \emph{@USER I know it was bad but I used to love it}
	\item FP: \emph{men who voted yes are childish, sorry are you 17??Men, would you have a problem if a girl said if she’s not receiving head she’s not giving head?}
	\item FN:  @USER It’s as if every single one of her supporters are just as stupid as her.  Wehdone..
	\item FN: I’m gonna say he doesn’t really. You should check the zip code demographics of his various properties Only liars could deny that Al Sharpton hates white people
	\item FN:  @USER Fuck the chargers
	\item FN: Last night I watched the Democrats throwing shit up against the wall, and none of it stuck.
\end{enumerate}
\end{tcolorbox}
}
\caption{False positive (FP) and false negative (FN) examples of our UHH-LT Task A submission.}
  \label{fig:examples}
\end{figure}

\section{Conclusion}

After last year's SemEval shared task on offensive language detection was already dominated by the then newly published BERT model, for the year 2020 competition we were successful in fine-tuning BERT's successor models to create the best performing system. The predictions obtained from fine-tuning of the ALBERT model on the OLID dataset achieved 92.25 \% macro F1-score as the best overall result (including our post-submission experiments) on the official test set of the Shared Task A for the English language.
We also found the `weak labels' distributed along with 9~million tweets by the shared task organizers not useful for training our classifiers. However, the tweets themselves provided useful in-domain data for unsupervised pre-training.
With 92.04 \% macro-F1, our predictions based on further language model pre-training of the RoBERTa model on ca. 440.000 tweets, before fine-tuning on last year's OLID dataset achieved the first rank in the official SemEval competition for sub-task A, and also high ranks (6, and 3) for the other two sub-tasks. We conclude that further pre-training of a transformer model with in-domain data is useful for offensive language detection. However, for tasks B and C, our models are clearly biased towards the majority class resulting in somewhat lower ranks. Hence, taking the high class imbalance of the OLID dataset better into account could further improve our results.

\section*{Acknowledgements}

The paper was supported by BWFG Hamburg (Germany) within the ``Forum 4.0'' project as part of the \textit{ahoi.digital} funding line, and the DFG project ``FAME'' (WI 4949/2-1, grant no. 406289255).

\bibliographystyle{coling}
\bibliography{coling2020}

\end{document}